\documentclass[%
reprint,
nofootinbib,
 amsmath,amssymb,
 aps,
]{revtex4-2}

\usepackage{physics}
\usepackage{braket}
\usepackage{romanbar}
\usepackage{graphicx}
\usepackage[caption=false]{subfig}
\usepackage{dcolumn}
\usepackage{tikz}
\usepackage{bm}
\usepackage{float}
\usepackage{chemformula}
\usepackage{xcolor}
\usepackage[symbol,hang,flushmargin]{footmisc}
\usepackage{bbold} 
\usepackage{caption}
\usepackage{float}

\begin{document}

\renewcommand\topfraction{.9}
\renewcommand\bottomfraction{.7}
\renewcommand\textfraction{.1}
\renewcommand\floatpagefraction{.95}

\preprint{APS/123-QED}

\title{
Imaging at the quantum limit with convolutional neural networks
}

\author{Andrew H. Proppe$^{1,2,3}$}
\author{Aaron Z. Goldberg$^{2}$}
\author{Guillaume Thekkadath$^{2}$}
\author{Noah Lupu-Gladstein$^{2}$}
\author{Kyle M. Jordan$^{3}$}
\author{Philip J. Bustard$^{2}$}
\author{Frédéric Bouchard$^{2}$}
\author{Duncan England$^{2}$}
\author{Khabat Heshami$^{1,2}$}
\author{Jeff S. Lundeen$^{1,3}$}
\author{Benjamin J. Sussman$^{1,2}$}

\affiliation{
$^1$
Joint Center for Extreme Photonics, University of Ottawa and National Research Council of Canada, 100 Sussex Drive, Ottawa, Ontario, Canada K1A 0R6
\\$^2$ National Research Council of Canada, 100 Sussex Drive, Ottawa, Ontario K1A 0R6, Canada
\\$^3$ 
Department of Physics and Nexus for Quantum Technologies, University of Ottawa, 25 Templeton Street, Ottawa, Ontario, Canada K1N 6N5
}

\begin{abstract}

{
Deep neural networks have been shown to achieve exceptional performance for computer vision tasks like image recognition, segmentation, and reconstruction or denoising. Here, we evaluate the ultimate performance limits of deep convolutional neural network models for image reconstruction, by comparing them against the standard quantum limit set by shot-noise and the Heisenberg limit on precision. We train U-Net models on images of natural objects illuminated with coherent states of light, and find that the average mean-squared error of the reconstructions can surpass the standard quantum limit, and in some cases reaches the Heisenberg limit. Further, we train models on well-parameterized images for which we can calculate the quantum Cramér-Rao bound to determine the minimum possible measurable variance of an estimated parameter for a given probe state. We find the mean-squared error of the model predictions reaches these bounds calculated for the parameters, across a variety of parameterized images. These results suggest that deep convolutional neural networks can learn to become the optimal estimators allowed by the laws of physics, performing parameter estimation and image reconstruction at the ultimate possible limits of precision for the case of classical illumination of the object.
}
\end{abstract}

\maketitle


\section{Introduction}
\label{section:introduction}

\begin{figure*}[t!]
    \centering
    \includegraphics[width=0.96\textwidth]{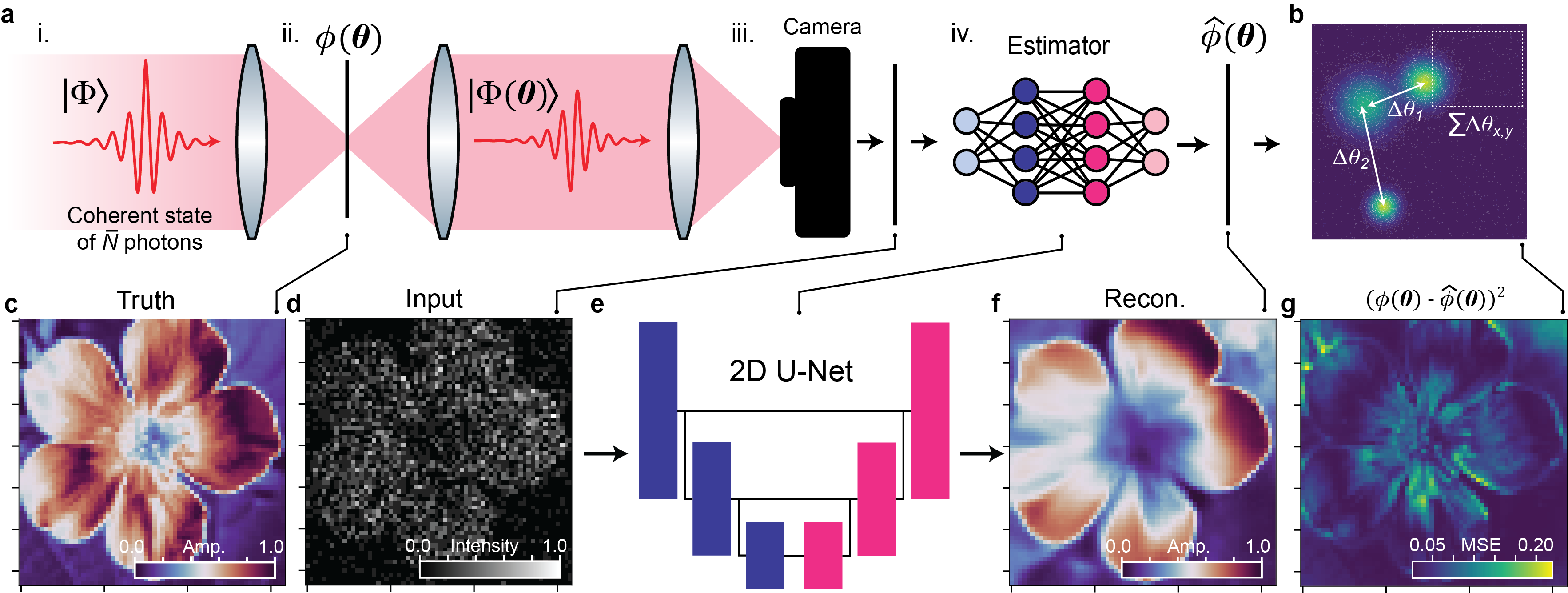}
    \caption{
    A general quantum photonic metrology scheme, and our image reconstruction task (a) Optical setup for measuring images and estimating parameters $\theta$. Here, we consider the estimation of the transmittance of an object, $\phi(\theta)$. In the framework of quantum metrology, the probe state $\ket{\Phi}$ is prepared (i) and then interacts with the system $\phi(\theta)$ parameterized by $\theta$ (ii). $\ket{\Phi(\theta}$ is measured (iii) and an estimator extracts information about $\theta$ (iv). (b) An estimated image from the average of multiple measurements which yields uncertainties $\Delta\theta_i$, which can be e.g. the distance between sources, or for full image reconstruction of un-parameterized objects, the variance of each predicted pixel. (c) Normalized true image $\phi$, (d) example input frame with units of normalized intensity, (e) schematic of the U-Net architecture, (f) reconstructed (estimated) image $\hat{\phi}$, and (g) the MSE residuals between the $\phi$ and $\hat{\phi}$.
    }  
    \label{fig:fig1}
\end{figure*}

Deep learning and artificial neural networks (NNs) have emerged as powerful and versatile data-driven algorithms, achieving state-of-the-art performance across diverse domains of science and engineering. NNs learn to extract meaningful patterns from large and complex datasets, leading to promising advances across diverse domains like natural language processing, audio and image analysis and generation, and materials discovery. The field of computer vision has benefited immensely from deep learning: tasks such as object classification \cite{resnets, ImageNet}, detection \cite{obj_detection_DL}, and segmentation \cite{UNet_segmentation} have seen enormous progress in recent years, driven by advancements in both algorithmic techniques and access to large-scale datasets like ImageNet \cite{ImageNet}. 

One particularly promising application of deep learning in computer vision and sensing is image reconstruction and signal de-noising. Image reconstruction involves restoring images from incomplete or noisy data---a task crucial in areas like medical imaging \cite{lequyer_fast_2022}, microscopy \cite{wang_deep_2020}, and remote sensing (e.g. satellite photography) \cite{proppe_PRUNe}. Architectures such as autoencoders \cite{proppe_adversarial_2022-1, proppe_PRL, raman_denoise}, U-Nets \cite{mao_symskip}, and generative adversarial networks \cite{ahmed_quantum_2021} have been highly effective in recovering fine details, even from heavily degraded or noisy inputs. Given these advances and the state-of-the-art performance by deep NNs for a large variety of computer vision tasks, a natural questions arises: for noisy inputs that create uncertainty in a model's prediction, what is the lowest possible uncertainty or variance bounded by the laws of physics for these tasks? And how close are NN predictions to those limits? Below, we will show that convolutional NNs perform image reconstruction with precisions that reach fundamental bounds set by quantum information theory.

In a typical imaging scenario, an object is illuminated by a beam of light--the probe--containing on average $N$ photons. These photons are detected on a camera to determine the object's transmittance or reflectance (Fig. 1a). For classical states of light (or classical-like coherent states such as from a laser pulse), the number of detected photons fluctuates due to the photon statistics. This leads to shot noise, a fundamental source of uncertainty that becomes significant at low illumination levels, and manifests as random variations in the detected pixel intensities. Shot-noise can be overcome using quantum states of light, but here we focus on the much more common and practical scenario of probing systems with coherent states of light.

It is often of interest to determine parameters from the measured image, e.g. the height or width of nanostructured surfaces \cite{Hermosa_nanostep}; or the distance between two light sources \cite{aaron_superresolution, Bonsma-Fisher_2019, Tsang2016}; or the intensity at each pixel for an entire image \cite{proppe_PRUNe} (Fig. 1b). If we tasked a NN with estimating these parameter values from a measured image, how does the model's estimate compare with fundamental precision bounds? 

Metrology, the science of measurement, aims to determine parameters of physical systems with the highest possible precision using finite resources, such as the number of photons illuminating an object and detected on a camera \cite{Sciarrino_quant_metro}. Quantum metrology describes the probe, and the effect of the physical system under investigation on the probe, using quantum theory (Fig. 1a): in this framework, a probe state $\ket{\Phi}$ is prepared, allowed to interact with a system $\phi(\theta)$ parameterized by $\theta$, and then $\ket{\Phi(\theta)}$ is measured to extract information about $\phi(\theta)$ (Fig. 1b). The measurements are given to an estimator, which is a mathematical function or algorithm that processes the measurement data to infer the value of one or more unknown parameters of interest as $\hat{\phi}$. A central question is how the precision---quantified by the variance of the parameter estimation---scales with the number of resources (here, photons), ${N}$.

The tools of quantum metrology allow us to derive and quantify fundamental precision limits of such a parameter estimation scheme using the quantum Fisher information to calculate the quantum Cramér-Rao bound (QCRB). This bound determines the minimum possible variance for an estimated parameter \cite{Villegas_PRL}, and is agnostic to the measurement setup used, depending only on the quantum probe state itself, $\ket{\Phi}$: it represents the ultimate limit of precision for a given set of parameters, $\theta$, and cannot be exceeded. For example, the QCRB for the aforementioned classical probe states is the shot-noise limit, also called the standard quantum limit (SQL), where the variance decreases as $1/N$. Employing nonclassical probes such as entangled photon pairs can give an advantage in measurement precision, achieving the Heisenberg limit (HL), where the variance scales as $1/N^2$ \cite{Maccone_metrology}.

Finding optimal estimators for a given probe state, by designing algorithms that reach these precision bounds in practical settings, remains a key challenge in quantum metrology. It is often highly non-trivial, if not impossible, to determine such an estimator: for example, how would one design an estimator when there is no clear parametrization at all, such as in the pixel-wise image reconstruction of a natural object? For this real-world metrology task, which we explore here, there is no known universal algorithm for constructing optimal estimators. Consequently, most metrological tasks carried out in experimental labs likely operate far from the QCRB, increasing measurement acquisition times and potentially preventing the extraction of meaningful patterns from data that are obscured by noise \cite{proppe_PRL}.

This absence of a universal approach for finding optimal estimators contrasts with the universal approximation theorem of NNs \cite{augustine2024surveyuniversalapproximationtheorems, RNN_universal}. This makes NNs suitable to become estimators for possibly any metrological task, parameterized or not, given enough training data. We asked earlier: how far are the minimum uncertainties of NN predictions from fundamental bounds determined from quantum metrology? We may rephrase this to ask instead: are deep NN models capable of learning to become the optimal estimator of a given metrological task?

\section{Reconstructions of natural objects}
\label{section:natural_images}

\begin{figure*}[t!]
    \centering
    \includegraphics[width=0.86\textwidth]{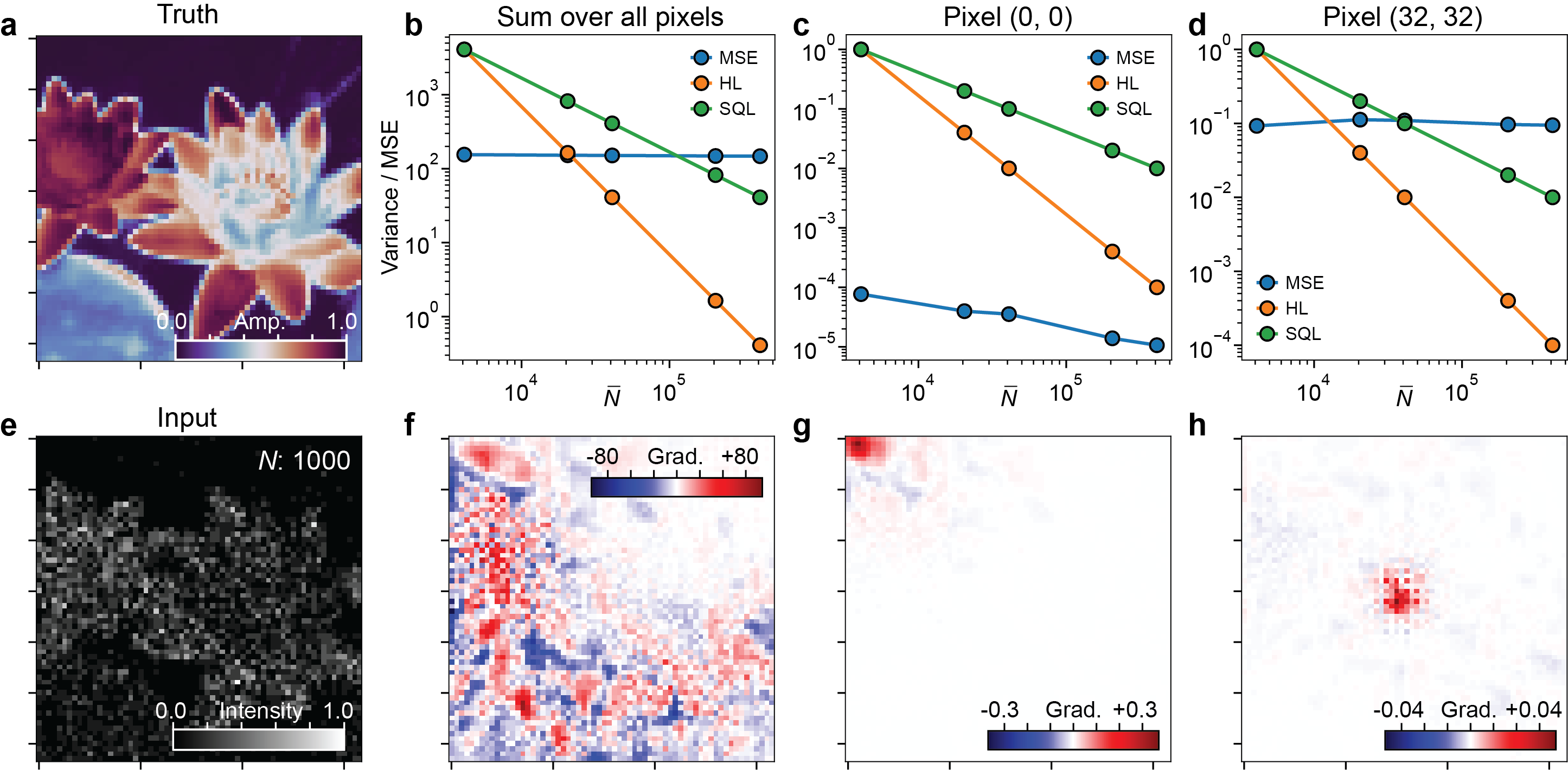}
    \caption{
    (a) Normalized true image $\phi$ used in this sample. (b) Average MSE, SQL variance, and HL variance versus $\bar{N}$ summed across all pixels, (c) the same for the top left pixel at (0, 0), and (d) for the centre pixel at (32, 32). (e) Example of an input image with $\bar{N}$ = 1000. (f) the backpropgated gradients for all pixels in the reconstructed image, (g) the gradients for the top left pixel at (0, 0) and (h) for the centre pixel at (32, 32).
    }
    \label{fig:fig2}
\end{figure*}

We begin our investigation by examining the reconstruction of intensity images $\phi(\theta)$ of natural objects, which lack explicit parametrization. In such a case, we treat the pixels as uncorrelated, and the parameters of $\theta$ we are trying to estimate are simply the intensities at each pixel. In our simulations, the illumination beam is a coherent state $\ket{\Phi}$ with mean photon number of $\bar{N}$ and complex-valued spatial profile $\alpha(x, y)$, which for simplicity we assume to be uniform across the image (i.e. $\abs{\alpha(x, y)}^2 = 1$).

Similar to ref. \cite{proppe_PRUNe}, we use images of flowers from the Oxford102 flowers dataset \cite{oxford102}, which serve as a diverse set of realistic and complex images with features at different spatial scales (Fig. \ref{fig:fig1}a). The model inputs are frames of simulated `measurements' of $\phi(\theta)$ with mean photon number $\bar{N}$ and pixel-wise values of $N(x,y)$. We introduce shot-noise in these frames that reflects the photon statistics of our coherent probe states by Poisson sampling at each pixel using $N(x,y)$ as the photon number expectation value, resulting in images as in Fig. \ref{fig:fig1}d. To train the model across a wide range of signal levels, $\bar{N}$ is randomly chosen from a uniform distribution between 100 - 10000. These levels of illumination may arise in scenarios where the light intensity is restricted due to diffuse reflection from an object, when probing photosensitive biological and molecular samples, or when image frames must acquired quickly to resolve dynamics. 

Our model is a 2D U-Net model adapted from ref. \cite{proppe_PRUNe} (Fig. 1e). This model takes as an input a single `measured' frame of size 64$\times$64, and outputs a normalized estimate, $\hat{\phi}(\theta)$, of the true image $\phi(\theta)$. Training details are given in the Supplementary Material. In the SQL and HL, the variance of the estimated parameter---here the pixel-wise reconstructed intensity---scales as $1/N(x,y)$ and $1/N^2(x,y)$, respectively. Our first task is to evaluate the model performance versus the variances from  the SQL and HL by comparing these variance limits with the pixel-wise mean-squared error (MSE) between the true $\phi(\theta)$ and reconstructed $\hat{\phi}(\theta)$ images, $(\phi_i(\theta) - \hat{\phi}_i(\theta))^2$.

MSE is a common loss function for training NN models, but in our context it has further significance: the MSE combines both the variance and squared bias in the estimation, and so is always larger (or equal to, if unbiased) than the error due to the variance alone. Since the SQL or HL reflect only the variance of the estimation, an MSE smaller than these limits indicates a superior performance both in terms of accuracy and precision. NNs are typically biased estimators due to their architectural and training constraints, which embed assumptions about the data and limit the flexibility of the model. This bias counteracts the model variance to prevent overfitting, enabling NNs to generalize effectively to unseen data. Using MSE as our metric accounts for the model bias, allowing for a fair comparison with SQL and HL variance limits.

To estimate the average model MSE, we use a set of frames from the same $\phi(\theta)$ and $\bar{N}$, but each with different random Poisson sampling. In this way, we simulate acquiring a set of individually measured images. We found that using model predictions from a set of 1000 frames was sufficient to obtain approximately normally distributed values at each pixel (Fig. \ref{fig:pixel_value_dist}). We take the MSE of each $\hat{\phi}(\theta)$ (Fig. \ref{fig:fig1}f) with respect to $\phi(\theta)$, to obtain the model's average MSE for a given $\phi(\theta)$ and $\bar{N}$ (Fig. \ref{fig:fig1}g).

Figure \ref{fig:fig2} shows a representative example for plots of the model MSE versus the SQL and HL variances at different values of $\bar{N}$ for i) the sum all pixels, ii) a pixel in the top left corner of the image, and iii) a pixel in the centre of the image. When summing over the images, we observe that for many values of $\bar{N}$, the model reconstructions have a total MSE below the total SQL and HL calculated by summing $1/{N(x,y)}$ and $1/N^2(x,y)$.

But this result is at odds with our understanding of fundamental precision limits for parameter estimation: estimates from these measurement images, simulated under classical illumination of a coherent state, should only be able to saturate the SQL, not exceed it. Further, the HL is normally exclusive to quantum states of light, and should never be reached (much less surpassed) with classical probes. An explanation for this may be that in our pixel-wise calculation of the total SQL and HL variances, we do not account for any correlations between neighboring pixel values - when in fact there are clearly correlations in these natural images, which our convolutional NN uses when making predictions. We visualize this with a Grad-CAM-like \cite{gradCAM} approach by inspecting the backpropagated gradients from the model outputs to the input image, which allows us to directly observe the weights of the input pixels used in the reconstruction. Fig. \ref{fig:fig2}f - h shows examples of this process, exemplifying how our model captures correlations across the input images and uses neighboring input pixels to predict the value at a single output pixel.

Moreover, there are clearly locations in the image where the variance limits stay constant from the uniform illumination beam, but which are easy for the NN to estimate: for example, the model predicts with very little error the featureless corner pixels in the background of the flower (Fig. \ref{fig:fig2}c), and thus appears to go well below the HL. The opposite is true for a pixel in the centre of the image (Fig. \ref{fig:fig2}d). How then should we make a fair comparison between our convolutional NN predictions, which implicitly capture correlations in the input image, and fundamental noise limits which must also reflect these correlations?

\section{Reconstructions of parameterized images}
\label{section:parameterized_images}

Such correlations may be captured in images where the pixel intensities are parameterized with some functional form. For example, if we know \textit{a priori} that our images are a simple linear function, we are only interested in the variance of our estimates for the slope and intercept parameters, since these can be used to describe every pixel intensity in the resulting image.

We now consider objects $\phi(\pmb{\theta})$ where $\pmb{\theta}$ is a vector of parameters that describe the intensities of the entire image, instead of individual pixel intensities as in the un-parameterized case. With this parametric model, we can directly calculate the QCRB for the elements of $\pmb{\theta}$ using the quantum Fisher information matrix, $\mathcal{F}$. The uncertainty for these parameters can then be used to obtain the pixel-wise uncertainty of the entire reconstructed image, which we can compare with the MSE of our NN predictions.

Following the methodology of Villegas et al. \cite{Villegas_PRL}, the elements $\mathcal{F}_{ij}$ for a coherent state as the probe beam $\ket{\Phi(\pmb{\theta})}$ after transmission through $\phi(\pmb{\theta})$ can be written as:




\begin{align}
    \mathcal{F}_{ij} = 4\bar{N}\int dxdy\abs{\alpha(x, y)}^2\left[\frac{\partial\phi}{\partial\theta_i}\right]\left[\frac{\partial\phi}{\partial\theta_j}\right]
    \label{eq:qfim}
\end{align}

\noindent
where $\alpha(x, y)$ is again uniform for all pixels. These matrix elements $\mathcal{F}_{ij}$ quantify how sensitively changes in $\pmb{\theta}$ affect the overall image. The QCRB states that the covariance matrix $\Sigma(\pmb{\theta})$ is lower bounded by the inverse of $\mathcal{F}$, giving the lowest possible variances that can be measured for $\pmb{\theta}$ using a coherent state with photon number $\bar{N}$.

\begin{figure*}[t!]
    \centering
    \includegraphics[width=0.96\textwidth]{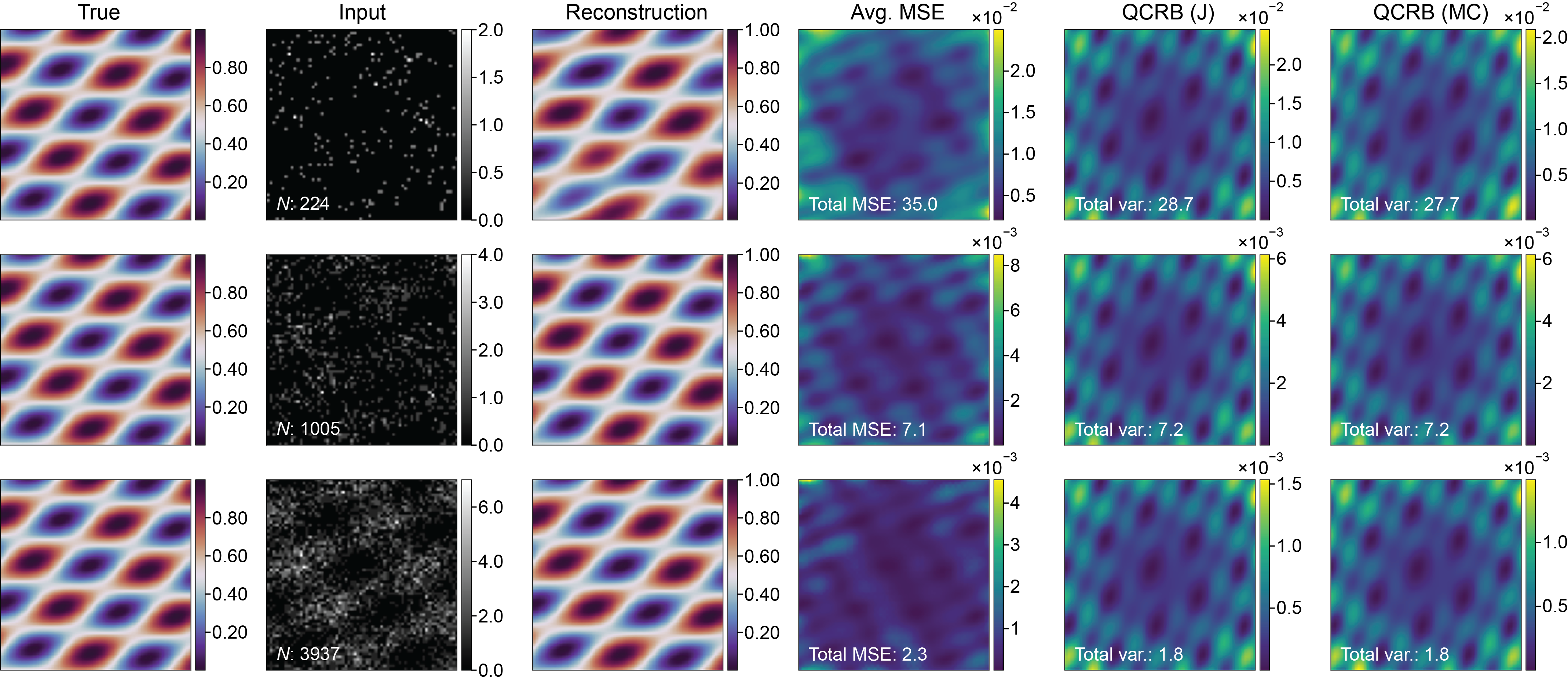}
    \caption{
    True image $\phi(\pmb{\theta})$; the model input; one of 1000 example reconstructions $\hat{\phi}(\pmb{\theta})$; the average mean-squared error (Avg. MSE) of the 1000 reconstructions; the pixel-wise QCRB variance calculated using the Jacobian and the covariance matrix (QCRB (J)); and using Monte Carlo sampling of the covariance matrix (QCRB (MC)). The three rows correspond to signal levels of $\bar{N}$ = ~100, 1000, and 10000. The sum of the photon counts for each input image are shown as insets, as well as the total MSE and QCRB variances.
    }
    \label{fig:fig3}
\end{figure*}

The domain of parameterized images we use for our analysis is of superposed linear and radial sinusoids, where we can control the amplitude, frequencies, angles, and phases of each component. This allows us to generate parameterized images that could be relevant in real imaging scenarios, while also tuning the number of elements of parameter vector $\pmb{\theta}$ between 3 and 11. We examine five different combinations: single, double, and triple linear sinusoids; a radial plus linear sinusoid; and a double radial sinusoid. For example, the equation below is used to generate two superposed linear sinusoids:

\begin{align}
    \phi(x, y) = \sum_{i=1,2}\ a_i\sin{(\omega_i(x\cos{\beta_i}+y\sin{\beta_i}+\varphi_i))}
    \label{eq:sinusoids}
\end{align}

\noindent
where $a_i$ is the amplitude, $\omega_i$ the frequency, 
$\beta_i$ the angle, and $\varphi_i$ the phase of sinusoid $i$. We set $a_2 = 1 - a_1$, and so $\theta = \left\{a_1, \omega_1, \beta_1, \delta_1, \omega_2, \beta_2, \delta_2\right\}$ giving a total of 7 free parameters for this image function.

We calculate $\mathcal{F}_{ij}$ for these parameters scaled by $\bar{N}$ numerically using the Python package SymPy, and invert this matrix to obtain the minimum possible $\Sigma(\pmb{\theta})$ as bounded by the QCRB. Since the NN does not estimate the parameters directly but reconstructs the entire image, we obtain pixel-wise variances by propagating the QCRB for each parameter into the final image using the Jacobian of $\pmb{\theta}$ and $\Sigma(\pmb{\theta})$. We also perform Monte Carlo sampling from $\Sigma(\pmb{\theta})$ to verify our calculations of the total variance in the generated image. See Supplementary Materials for details of both these methods.

We train the same 2D U-Net model to perform image reconstruction for each class of parameterized images, with varying levels of simulated shot-noise. Each dataset consists of 64,000 sinusoid images with $\bar{N} \simeq$ 40.96 - 4096 photons each (corresponding to ~0.01 - 1.0 photons-per-pixel), which undergo random Poisson sampling in each training epoch. Training loss curves are given in Fig. \ref{fig:sinusoid_losses}. To compare individual image reconstructions from the trained model with the QCRB, the same $\phi(\pmb{\theta})$ is used to generate 1000 model inputs at fixed $\bar{N}$ with different random Poisson sampling. Figure \ref{fig:fig3} shows a representative example for double linear sinusoids of true images $\phi(\pmb{\theta})$, model inputs, and reconstructions $\hat{\phi}(\pmb{\theta})$ for three different values of $\bar{N}$: 250, 1000, and 4000. Note that these images were generated on-the-fly for testing, and do not come from either the test or validation datasets, averting risks of memorized predictions. From the single input frame, we observe the 2D U-Net model returns accurate reconstructions at all signal levels. We further show the MSE image averaged over the 1000 inputs, as well as the maps of the pixel-wise QCRB variance calculated using the Jacobian and with Monte Carlo sampling. We observe that for the three values of $\bar{N}$, the total MSE fluctuates closely around the total variance from the QCRB, decreasing with increasing $\bar{N}$.

\begin{figure*}[t!]
    \centering
    \includegraphics[width=0.96\textwidth]{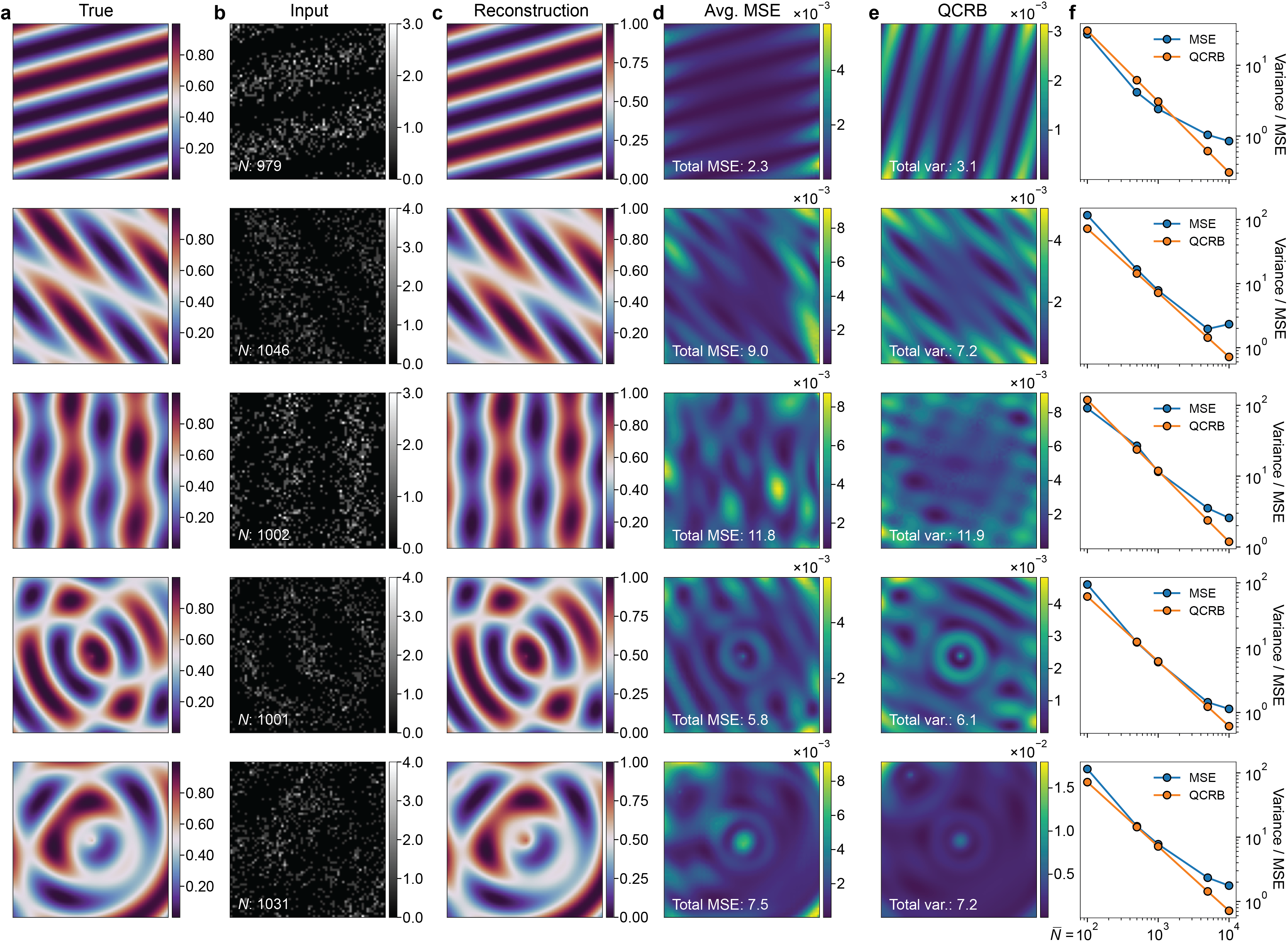}
    \caption{
    (a) True images, (b) model inputs, (c) reconstructions, (d) avg. MSE, (e) QCRB variances for (from top to bottom): a single; double; and triple linear sinusoid; a radial and linear sinusoid; and a double radial sinusoid for $\bar{N} \simeq 1000$. (f) MSE and QCRB versus $\bar{N}$ for the different sinusoid patterns.
    }
    \label{fig:fig4}
\end{figure*}

Exemplary results for the other sinusoid functions are shown in Fig. \ref{fig:fig4}, with more visualized reconstructions for other values of $\bar{N}$ in Figs. \ref{fig:linear_sinusoids} - \ref{fig:double_radial}. For the different sinusoid combinations, we can see that the total variance from the QCRB depends on the function complexity itself: for example, the single linear sinusoid has only 3 free parameters with total QCRB of 3.1, whereas the triple linear sinusoid, which has 11 parameters, has a total QCRB of 11.9. We further observe that the MSE of the U-Net reconstructions also depends on the complexity of the sinusoid function: it consistently follows the QCRB for all image types, reflecting the fact that minimum NN uncertainty also increases with function complexity. This may suggest that the optimal NN performance is indeed tied to the intrinsic variance of the parametric images, and each model learns to become the optimal estimator for each image domain.


We also show the MSE and QCRB as a function of $\bar{N}$ in \ref{fig:fig4}f, showing how the two metrics stay similar as $\bar{N}$ increases. The last two values of $\bar{N}$ plotted here, 5000 and 10000, are outside of the range used for training ($\bar{N}$ = 40.96 - 4096), and we see the model MSE begins to diverge from the QCRB. Perhaps unsurprisingly, we posit that the model learns to become the optimal estimator only for the domain of images and signal levels it was trained on, and does not extend this `ideal' performance outside of this domain. Fig. \ref{fig:different_training_N} shows plots of the MSE and QCRB versus $\bar{N}$ for models trained with $\bar{N}$ ranges of 40 - 4000, 40 - 40000, and 400 - 40000. We see that extending the range of $\bar{N}$ to 40000 increases MSE at lower $\bar{N}$ and decreases it at higher $\bar{N}$, consistent with this hypothesis.

We also plot the MSE and QCRB calculated when assuming the pixels are uncorrelated, as for the un-parameterized images of flowers (i.e. $\pmb{\theta}$ is the intensity at each pixel, and the QCRB = $\sum_{x,y} 1/N(x,y)$), shown in Fig. \ref{fig:mse_vs_SQL_uncorrelatd}. The MSE appears orders of magnitude below the SQL, again demonstrating how assuming uncorrelated variances of the pixels greatly overestimates the fundamental error bound compared to using the quantum Fisher information matrix.


\section{Discussion}
\label{section:disc}

This fundamental result suggests the NN model allows us to estimate the parametric image in the ultimate limit of minimal variance for a given probe state. This observation may give an explanation as to why our model trained on images of flowers seemed to surpass the HL under the naive assumption of no correlations between neighboring pixels: perhaps if it was possible to use a parametric model for flowers instead, we would find that the NN reconstructions for such realistic objects would reach, rather than exceed, the SQL (i.e. the QCRB for a coherent state). Instead, we may cautiously extrapolate from our results from parameterized images to suggest that deep convolutional NNs, when trained on a specific domain of images and a specific range of photon numbers $\bar{N}$, learn to become the optimal estimators of those images, obviating the need for a known parametric model altogether.

The generalization of our results to other NN models remains an open question: is reaching the QCRB only achievable using these highly optimized U-Net architectures, or can other emerging image reconstruction models (e.g. vision transformers) also reach these limits? Is it a coincidence that our U-Net model seems to saturate the QCRB, or is this possibly a fundamental property of convolutional NNs? And would models trained on much larger amounts of data (e.g. a foundation model like ImageNet \cite{ImageNet}) also be able to reach - or even decisively surpass - these fundamental limits?

There is also the question of measurement resources, which here we consider to be the total number of photons used in each input image frame, $\bar{N}$. With 50 epochs of the 64,000 training images  an average $\bar{N}$ of 2000, a total of $6.4\times10^{9}$ photons are used in the model training. Do all or some of these photons need to be accounted for in our QCRB calculations, or can they be ignored considering the training is `offline' and the model parameters fixed for testing on individual images? Future studies addressing this question may benefit from examining the influence of the amount of training resources on the performance of simpler networks for parameter estimation tasks.

Finally, our results suggest that, in the limit of small mean photon number $\bar{N}$, classical precision limits in optical imaging may be reached using NN estimators. If a deep NN can be used to reach the SQL using classical states of light, then could it also be used to reach the HL limit using nonclassical states of light?


\section{Conclusion}
\label{section:Conclusions}

We have demonstrated that deep convolutional neural network models are capable of performing image reconstruction tasks in the ultimate quantum limit of minimal variance. By calculating the quantum Fisher information matrix for well-parameterized images of sinusoids, we find that the mean-squared error of our U-Net reconstructions can reach the minimum measurable variance for a given image. We posit that when trained on a specific domain of images, a deep convolutional neural network with sufficient capacity learns to become the optimal estimator for reconstructing an image given a limited number of measurement resources. 
Future studies will examine the generality of this capability for other neural network models, and determine how far these fundamental precision limits can be pushed by combining deep neural networks with nonclassical measurements.

\begin{acknowledgments}
We acknowledge the support of the Natural Sciences and Engineering Research Council of Canada (NSERC), Canada Research Chairs, the Transformative Quantum Technologies Canada First Excellence Research Fund (CFREF), and University of Ottawa-NRC Joint Centre for Extreme Quantum Photonics (JCEP) via the Quantum Sensors Challenge Program at the National Research Council of Canada. The NRC headquarters is located on the traditional unceded territory of the Algonquin Anishinaabe and Mohawk people.
\end{acknowledgments}

\section*{Code Availability}
Code for generating data and for training and testing models is available at \url{https://github.com/andrewhproppe/PhaseRetrievalNNs}.

\bibliographystyle{apsrev4-1} 
\bibliography{apssamp}

\onecolumngrid
\appendix
\pagebreak

\renewcommand{\thefigure}{S\arabic{figure}}
\setcounter{figure}{0}

\section{Supplementary materials}
\subsection{Model training and hyperparameters}

The deep learning stack was implemented using PyTorch \cite{torch} and PyTorch Lightning, with weight updates performed using the Adam optimizer \cite{adam_optimizer}; code and values for the data generation, hyperparameters, training, and evaluation pipeline can be found in the following Github repository: https://github.com/andrewhproppe/PhaseRetrievalNNs. For further information about the model architecture, see ref. \cite{proppe_PRUNe} for details about the ResNet blocks and the encoder (decoder) layers. Model evaluation and inference was performed on CPUs (2.6 GHz 6-Core Intel Core i7, 32GB RAM). Hyperparameters for the optimized 2D U-Net models are given below. The model learning rates were updated dynamically during training using the PyTorch functionality "Reduce Learning Rate On Plateau".

The objective function for our model training consists of three loss terms: mean-squared error (MSE), structural similarity (SSIM), and gradient difference loss (GDL). The MSE guides the model towards pixel-wise accurate reconstructions, but alone gives overly smooth images. We found previously that the SSIM and GDL loss terms were essential to predict smooth images, and also result in lower overall MSE. See ref. \cite{proppe_PRUNe} for more details.

\begin{table}[h]
    \centering
    \begin{tabular}{c c c}
        \toprule
        \textbf{Parameter} & \textbf{Value} \\
        Num. layers & 6 \\
        Conv. channels & 256 \\
        Kernels & 5, $3^a$ \\
        Downsample & 4 \\
        Dropout & 0.0 \\
        Activation & "GeLU" \\
        SSIM window size & 11 \\
        Initial learning rate & 5e-4 \\
        Weight decay & 1e-6 \\
    \end{tabular}
    \caption{Hyperparameters for U-Net models. $^a$: a kernel size of 5 was used in the first (last) layer of the encoder (decoder), and a kernel size of 3 was used for the remaining layers.}
    \label{tab:3D2D_unet_hparams}
\end{table}

\subsection{Datasets and data generation}

For the un-parameterized images of flowers, we trained our models on images of the flowers from the Oxford102 dataset \cite{oxford102}. Details for simulating noisy frames can be found in ref. \cite{proppe_PRUNe}.

For the parameterized images of sinusoids, we generated 64,000 images of different sinusoid combinations. We consider 5 different combinations: a single linear sinusoid, a double linear sinusoid, a triple linear sinusoid, a linear + radial sinusoid, and a double radial sinusoid. The single and triple linear sinusoids follow the same form as eq. \ref{eq:sinusoids}, with the sum changed to 1 or 3. The single, double, and triple linear sinusoids consist of 3, 7, and 12 parameters, respectively. The radial linear sinusoids used the form:

\begin{align}
    \phi(x, y) = 
    a\sin{(\omega_r\sqrt{x^{2}+y^{2}}+\varphi_r)}
    + (1 - a)\sin{(\omega_l(x\cos{\beta_l}+y\sin{\beta_l}+\varphi_l)}
    \label{eq:radial_linear}
\end{align}

\noindent where $a$ is the relative amplitude, $\omega_{r(l)}$, and $\varphi_{r(l)}$ are the frequency, and phase of the radial (linear) sinusoids, and $\beta_l$ is the angle of the linear component, for a total of 6 parameters. The double radial sinusoid is given by:

\begin{align}
    \phi(x, y) = 
    a_1\sin{(\omega_1\sqrt{x^{2}+y^{2}}+\varphi_1)}
    + (1 - a_1)\sin{(\omega_2\sqrt{(x-x_0)^{2}+(y-y_0)^{2}}+\varphi_2)}
    \label{eq:double_radial}
\end{align}

\noindent where the parameters are the same as the radial function above, but with the introduction of $x_0$ and $y_0$ as positional offsets for the second radial sinusoid for a total of 7 parameters.

The table below shows the bounds for the parameters for the generated functions:

\begin{table}[h]
    \centering
    \begin{tabular}{c c c}
        \toprule
        \textbf{Parameter} & \textbf{Lower bound}  & \textbf{Upper bound} \\
        $a$  & 0.0 & 1.0 \\
        $\omega$ & $0.5/n_{pix.}$ & $4.0/n_{pix.}$ \\
        $\varphi$  & $-\pi$ & $+\pi$ \\
        $\beta$ & $-\pi$ & $+\pi$ \\
        $x_0$ & $-n_{pix.}/2$ & $+n_{pix.}/2$\\
        $y_0$ & $-n_{pix.}/2$ & $+n_{pix.}/2$\\
    \end{tabular}
    \caption{Bounds for parameters that are randomly sampled to generate the sinusoid images. $n_{pix.}$ is the number of pixels used in the images, which in this work are $64\times64$ pixels.}
    \label{tab:nn_model_errors}
\end{table}

\clearpage

\subsection{Quantum Fisher information matrix calculations}

Here we demonstrate how the pixel-wise QCRB variances were calculated for the parameterized images.

First, we calculate the Jacobian, which is a vector of partial derivatives of the function $f$ with respect to the parameters \(\theta_1, \theta_2, \theta_3\):
\[
\mathbf{J}(x_1, x_2) = 
\begin{bmatrix}
\frac{\partial f}{\partial \theta_1} \\
\frac{\partial f}{\partial \theta_2} \\
\frac{\partial f}{\partial \theta_3}
\end{bmatrix}.
\]




The QFIM matrix is computed as:
\[
\text{QFI}_{ij} = 4 \cdot \int \frac{\partial f}{\partial \theta_i} \frac{\partial f}{\partial \theta_j} \, dx_1 dx_2,
\]

\noindent where \(\frac{\partial f}{\partial \theta_i}, \frac{\partial f}{\partial \theta_j}\) are the Jacobian components. Below are some example terms of the QFIM:
\begin{align*}
\text{QFI}_{11} &= 4 \cdot \int \left(\frac{\partial f}{\partial \theta_1}\right)^2 \, dx_1 dx_2, \\
\text{QFI}_{12} &= 4 \cdot \int \frac{\partial f}{\partial \theta_1} \cdot \frac{\partial f}{\partial \theta_2} \, dx_1 dx_2.
\end{align*}

The covariance lower bound is then obtained by inverting the QFIM:
\[
\Sigma = (\text{QFI})^{-1}.
\]

The variance of the image at each pixel \((x_1, x_2)\) is then calculated as:

\[
\text{Var}[f(x_1, x_2)] = \mathbf{J}(x_1, x_2)^T \Sigma \mathbf{J}(x_1, x_2),
\]

We also used an alternative Monte Carlo sampling approach to verify our calculations using $\mathbf{J}$ and $\mathbf{\Sigma}$. We draw $n_{\text{samples}}$ samples of $\pmb{\theta_s}$ from:
    \[
    \theta \sim \mathcal{N}(\pmb{\theta}, \Sigma).
    \]

\noindent where $\mathcal{N}$ is a multivariate normal distribution using the true parameter values $\pmb{\theta}$ and the covariance matrix $\Sigma$. For each sample $\pmb{\theta_s}$, we evaluate \(f(x_1, x_2; \theta_s)\) to generate an image. The variance at each pixel is then calculated as:
    \[
    \text{Var}[f(x_1, x_2)] = \frac{1}{n_{\text{samples}}} \sum_{s=1}^{n_{\text{samples}}} \left(f(x_1, x_2; \theta_s) - \bar{f}(x_1, x_2)\right)^2.
    \]

\clearpage

\subsection{Supporting figures}

\begin{figure*}[h]
    \centering
    \includegraphics[width=0.96\textwidth]{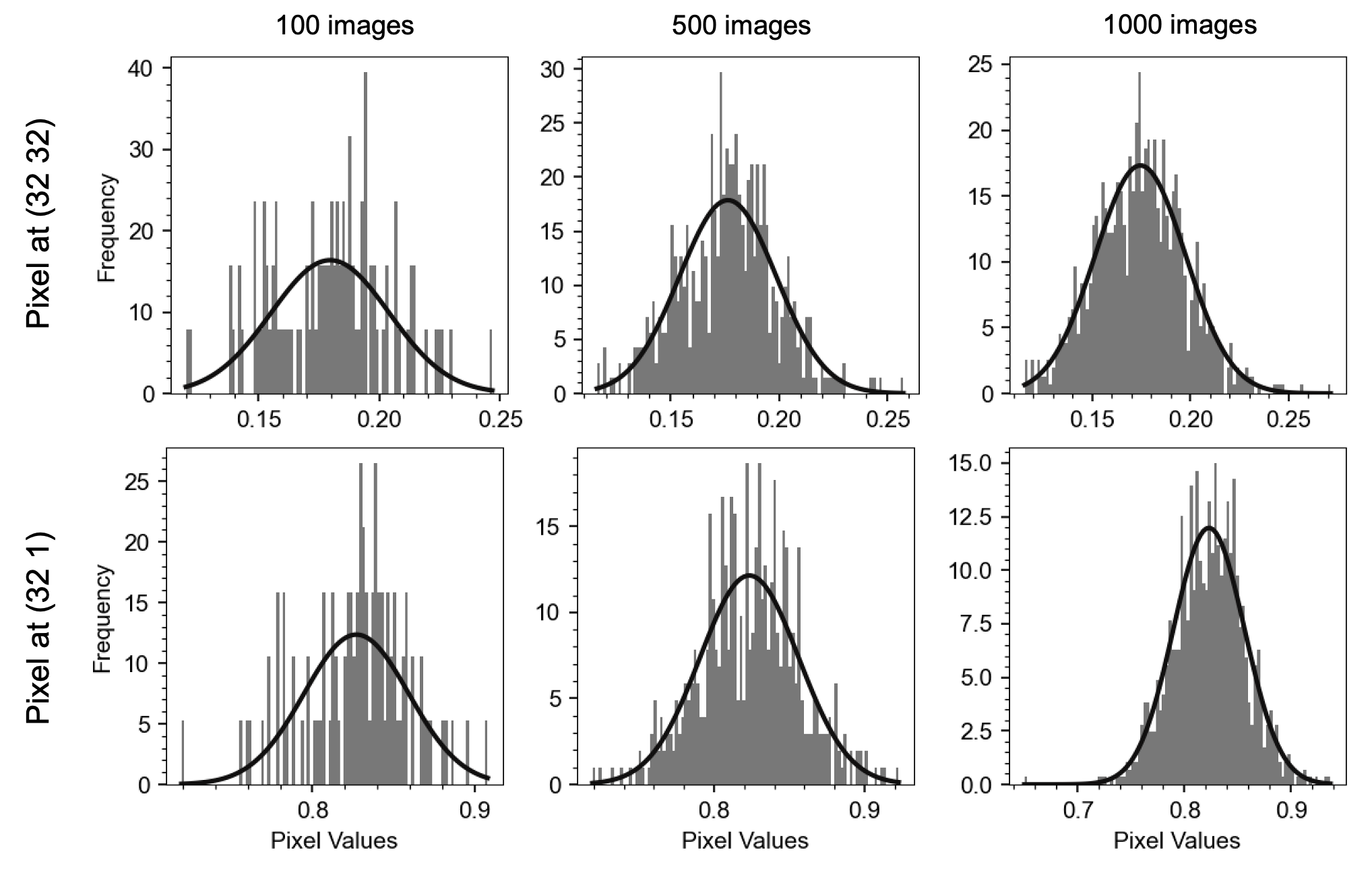}
    \caption{
    Histograms for the pixel values at pixel (32, 32) (top row) and (32, 1) (bottom row), for 100, 500, and 1000 images, with Gaussian fits to the distributions.
    }
    \label{fig:pixel_value_dist}
\end{figure*}

\begin{figure*}[h]
    \centering
    \includegraphics[width=0.96\textwidth]{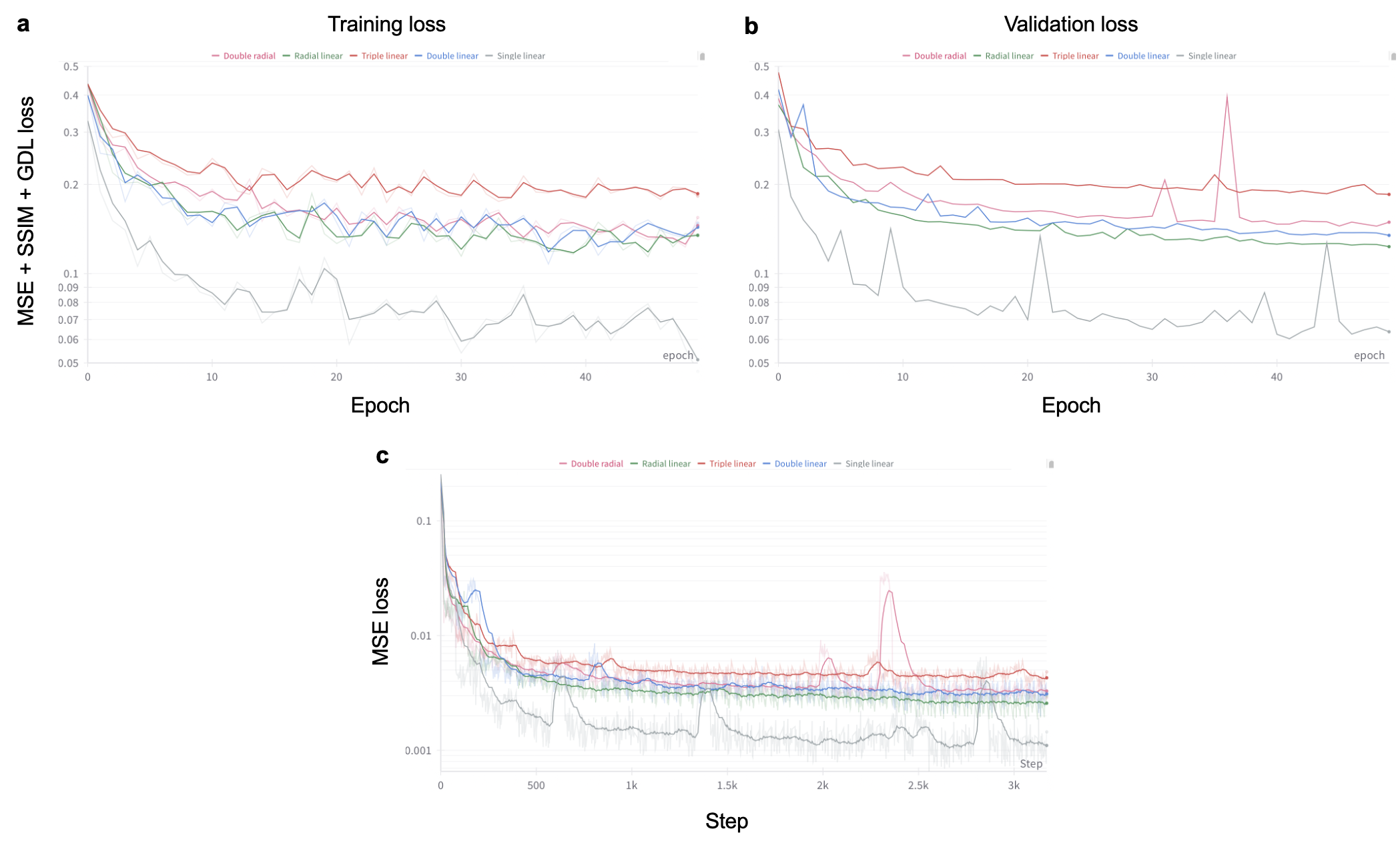}
    \caption{
    (a) Training loss, (b) validation loss, and (c) MSE loss (from the validation data) for the five 2D-to-2D U-Net models trained on images of sinusoids. Time-weighted averaging is applied to some of the curves for clarity. The training and validation losses are the sum of three loss terms: mean-squared error (MSE), structural similarity (SSIM), and gradient difference loss (GDL).
    }
    \label{fig:sinusoid_losses}
\end{figure*}

\begin{figure*}[h]
    \centering
    \includegraphics[width=0.96\textwidth]{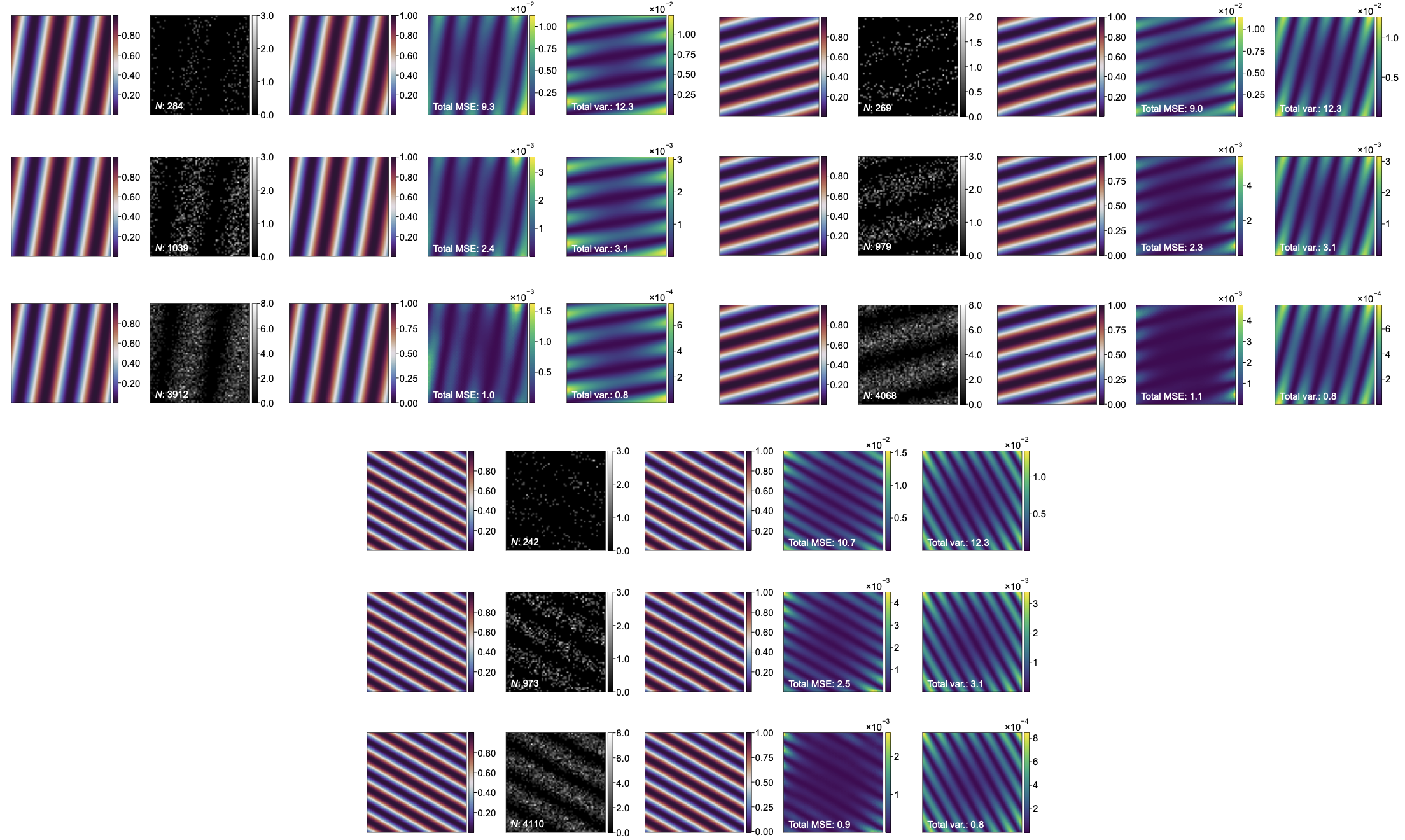}
    \caption{
    Three examples of a single linear sinusoid image, a model input, the model's reconstruction, the average MSE of 1000 reconstructions, and the image variance calculated using the Jacobian and covariance matrix.
    }
    \label{fig:linear_sinusoids}
\end{figure*}

\begin{figure*}[h]
    \centering
    \includegraphics[width=0.96\textwidth]{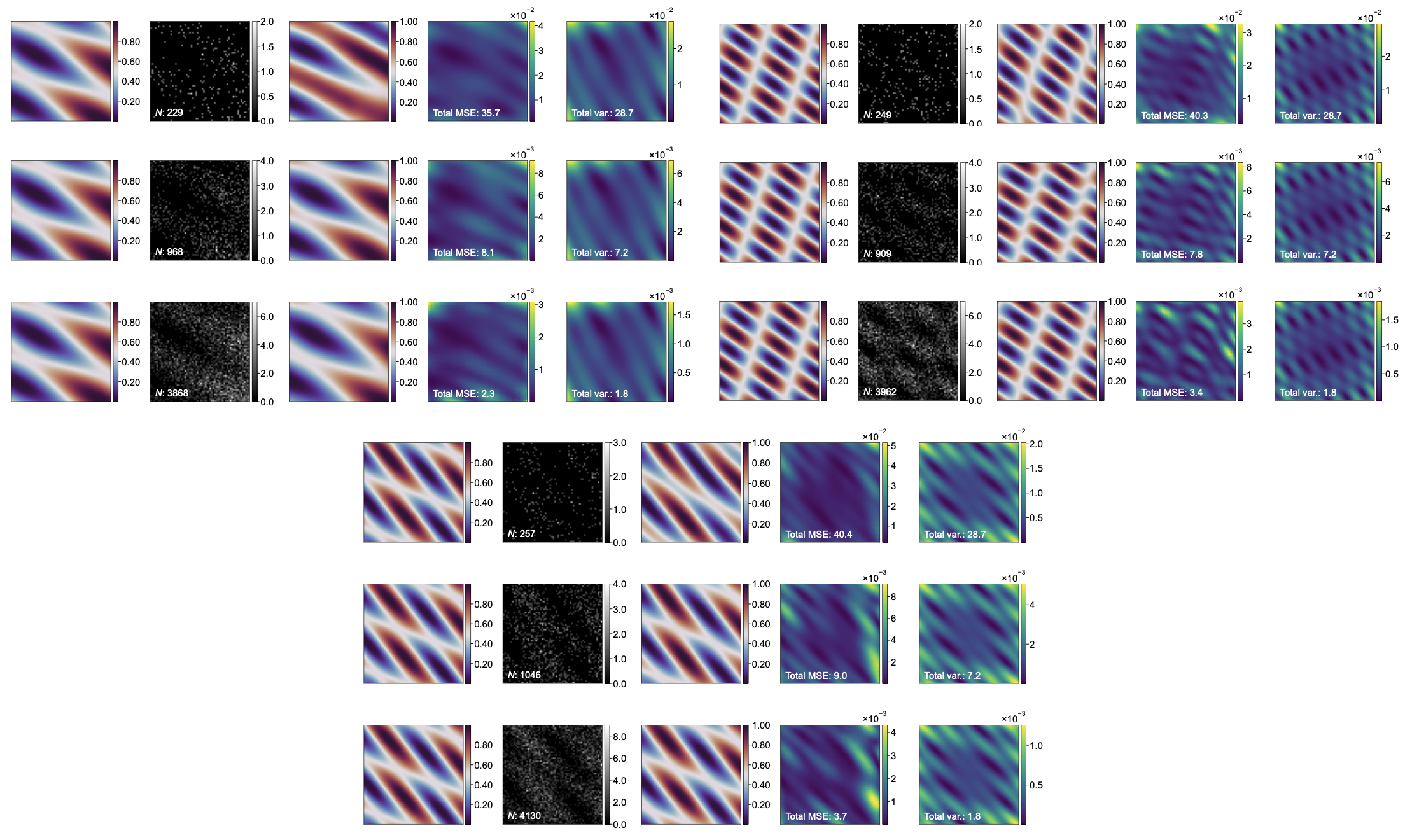}
    \caption{
    Three examples of a double linear sinusoid image, a model input, the model's reconstruction, the average MSE of 1000 reconstructions, and the image variance calculated using the Jacobian and covariance matrix.
    }
    \label{fig:double_sinusoids}
\end{figure*}

\begin{figure*}[h]
    \centering
    \includegraphics[width=0.96\textwidth]{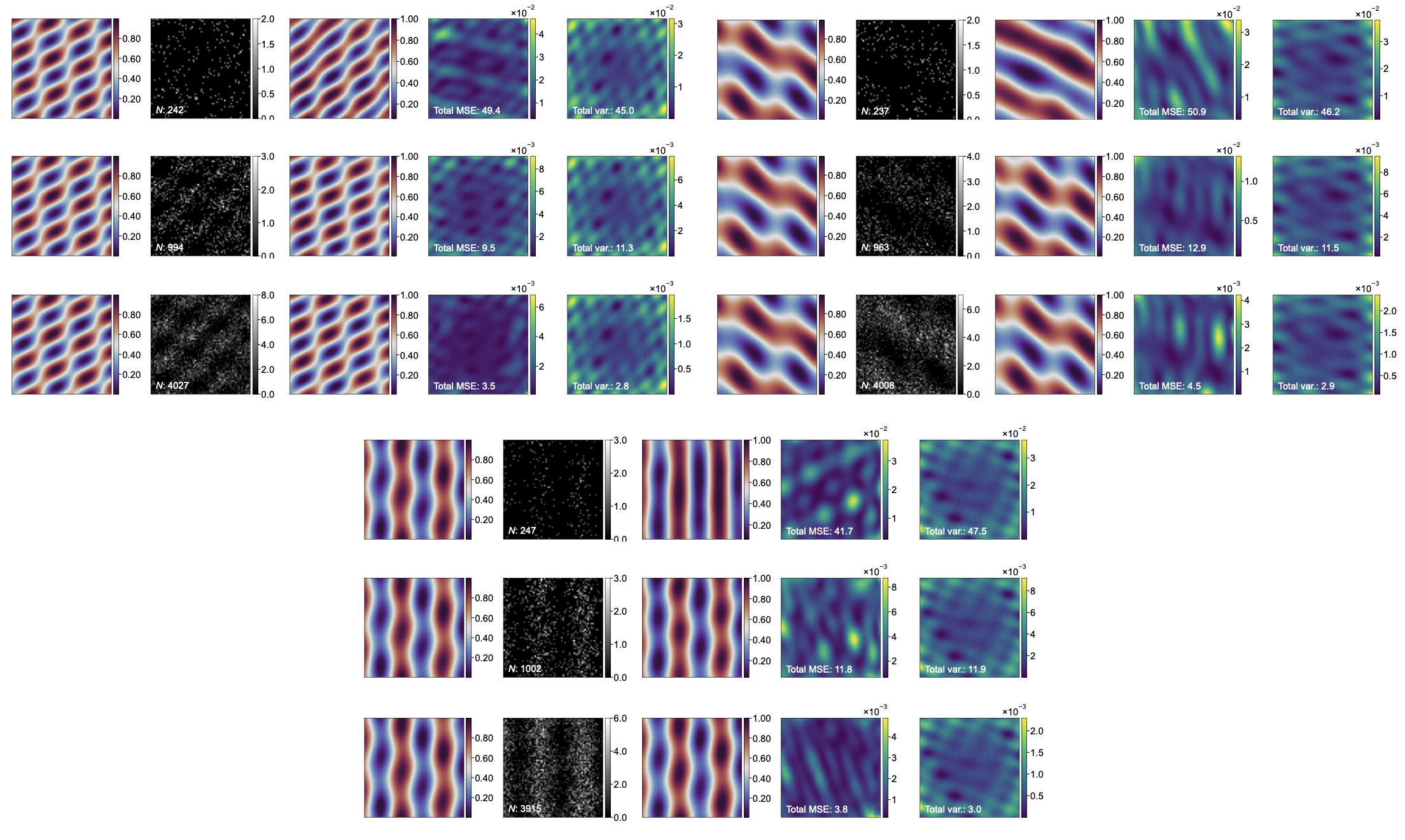}
    \caption{
    Three examples of a triple linear sinusoid image, a model input, the model's reconstruction, the average MSE of 1000 reconstructions, and the image variance calculated using the Jacobian and covariance matrix.
    }
    \label{fig:triple_sinusoids}
\end{figure*}

\begin{figure*}[h]
    \centering
    \includegraphics[width=0.96\textwidth]{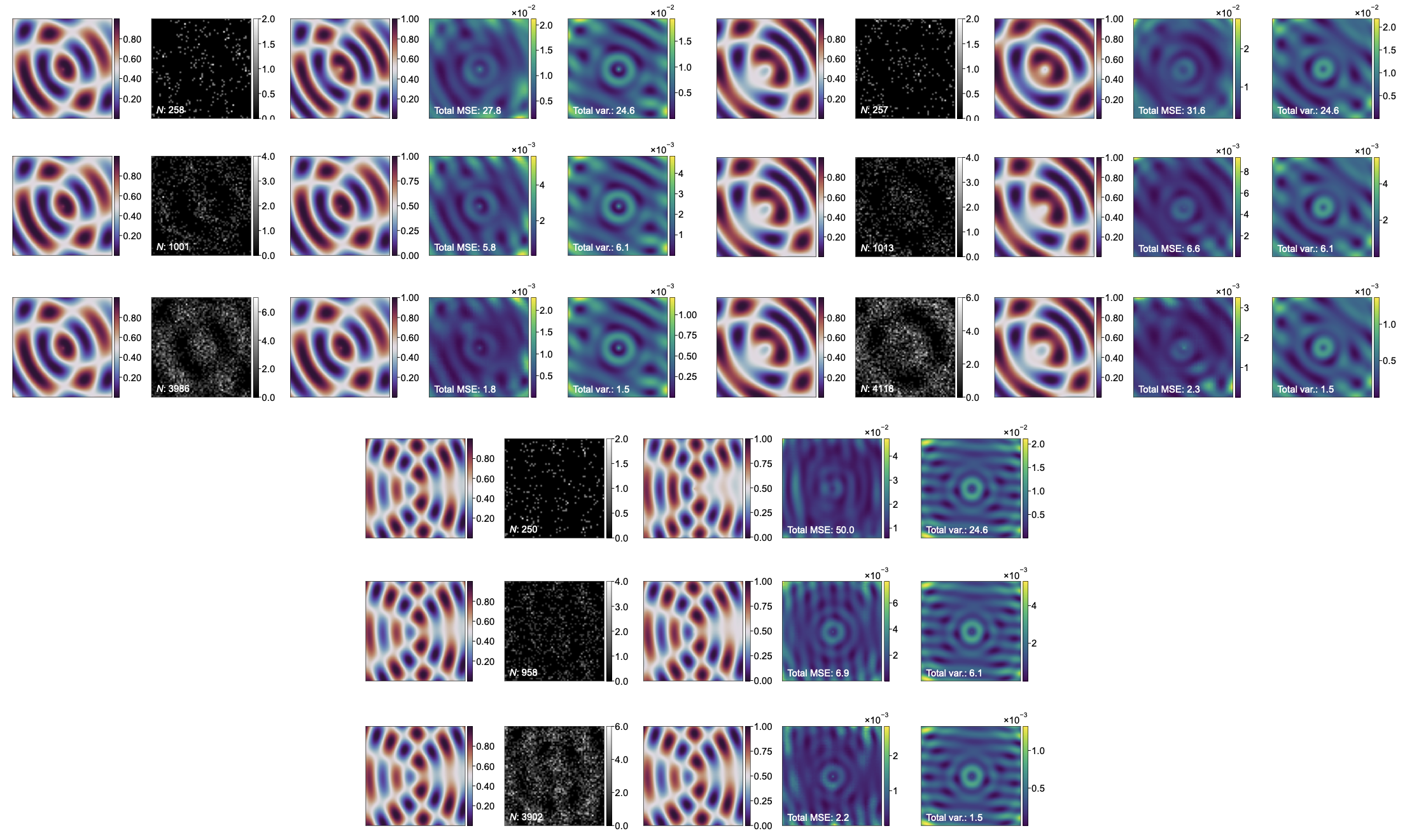}
    \caption{
    Three examples of a radial + linear sinusoid image, a model input, the model's reconstruction, the average MSE of 1000 reconstructions, and the image variance calculated using the Jacobian and covariance matrix.
    }
    \label{fig:radial_linear}
\end{figure*}

\begin{figure*}[h]
    \centering
    \includegraphics[width=0.96\textwidth]{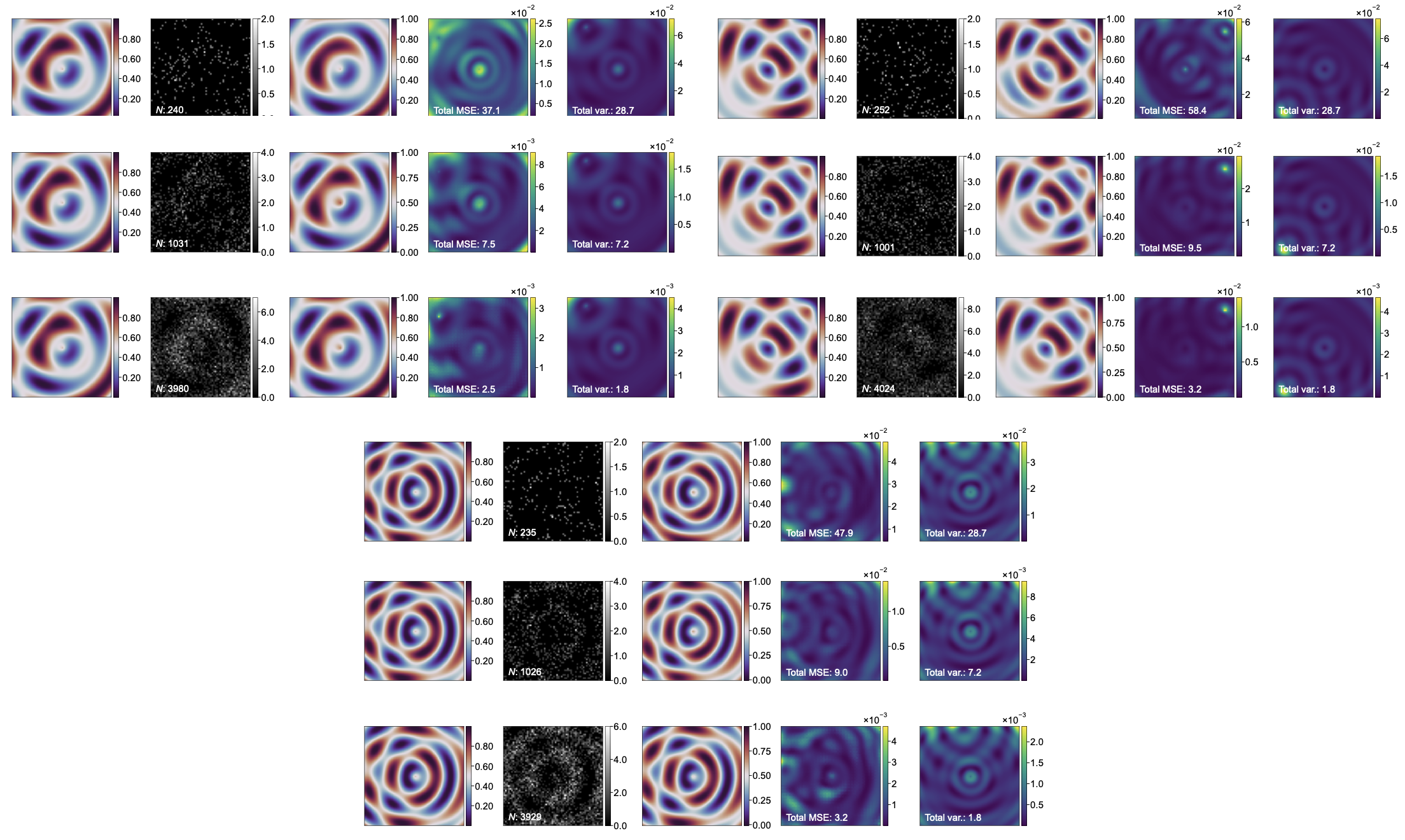}
    \caption{
    Three examples of a double radial sinusoid image, a model input, the model's reconstruction, the average MSE of 1000 reconstructions, and the image variance calculated using the Jacobian and covariance matrix.
    }
    \label{fig:double_radial}
\end{figure*}

\begin{figure*}[h]
    \centering
    \includegraphics[width=0.46\textwidth]{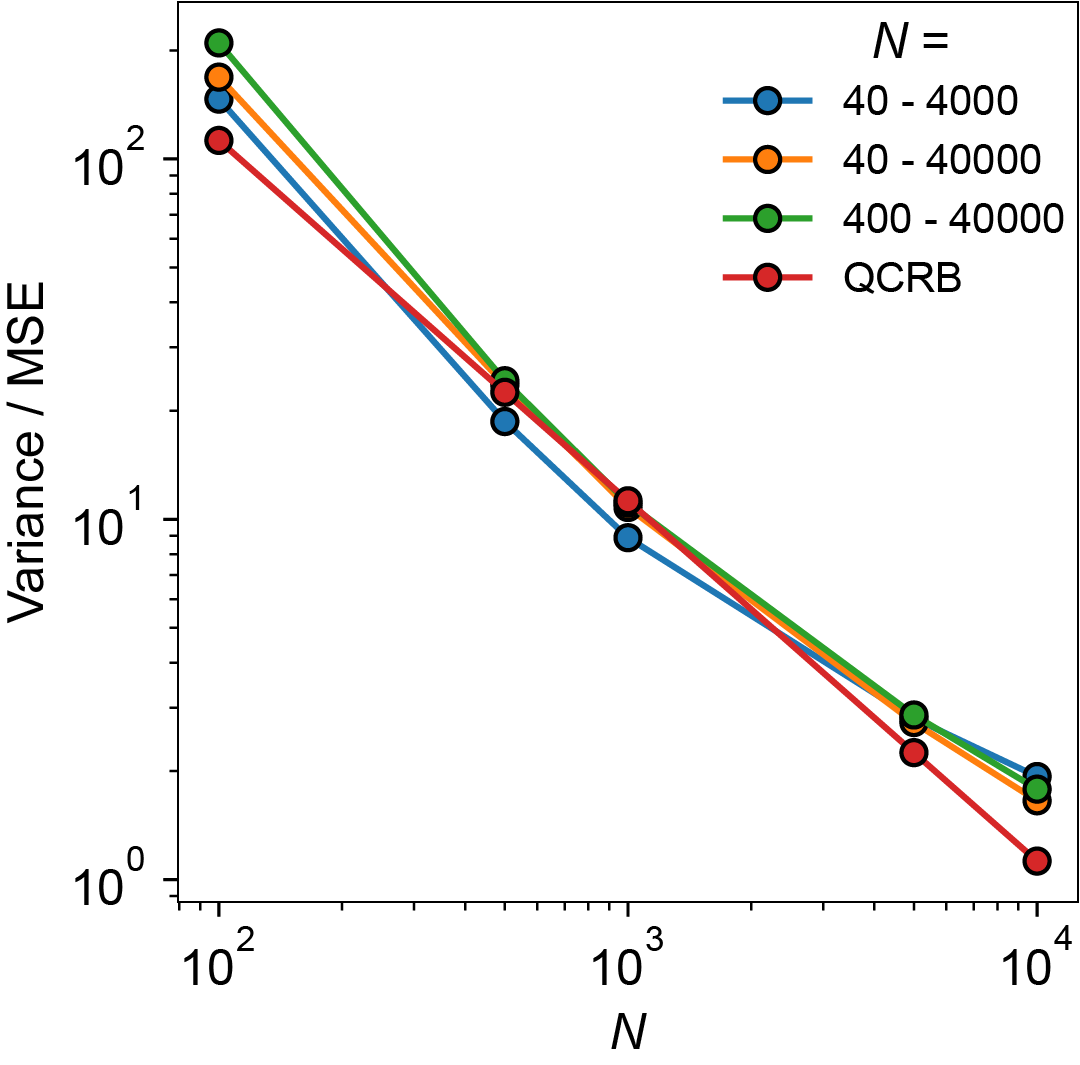}
    \caption{
    Model MSE and QCRB versus $\bar{N}$ for models trained on images of triple linear sinusoids with $\bar{N}$ ranges of 40 - 4000, 40 - 40000, and 400 - 40000 (corresponding to 0.01 - 1, 0.01 - 10, and 0.1 - 10 photons per pixel, respectively).
    }
    \label{fig:different_training_N}
\end{figure*}

\begin{figure*}[h]
    \centering
    \includegraphics[width=0.46\textwidth]{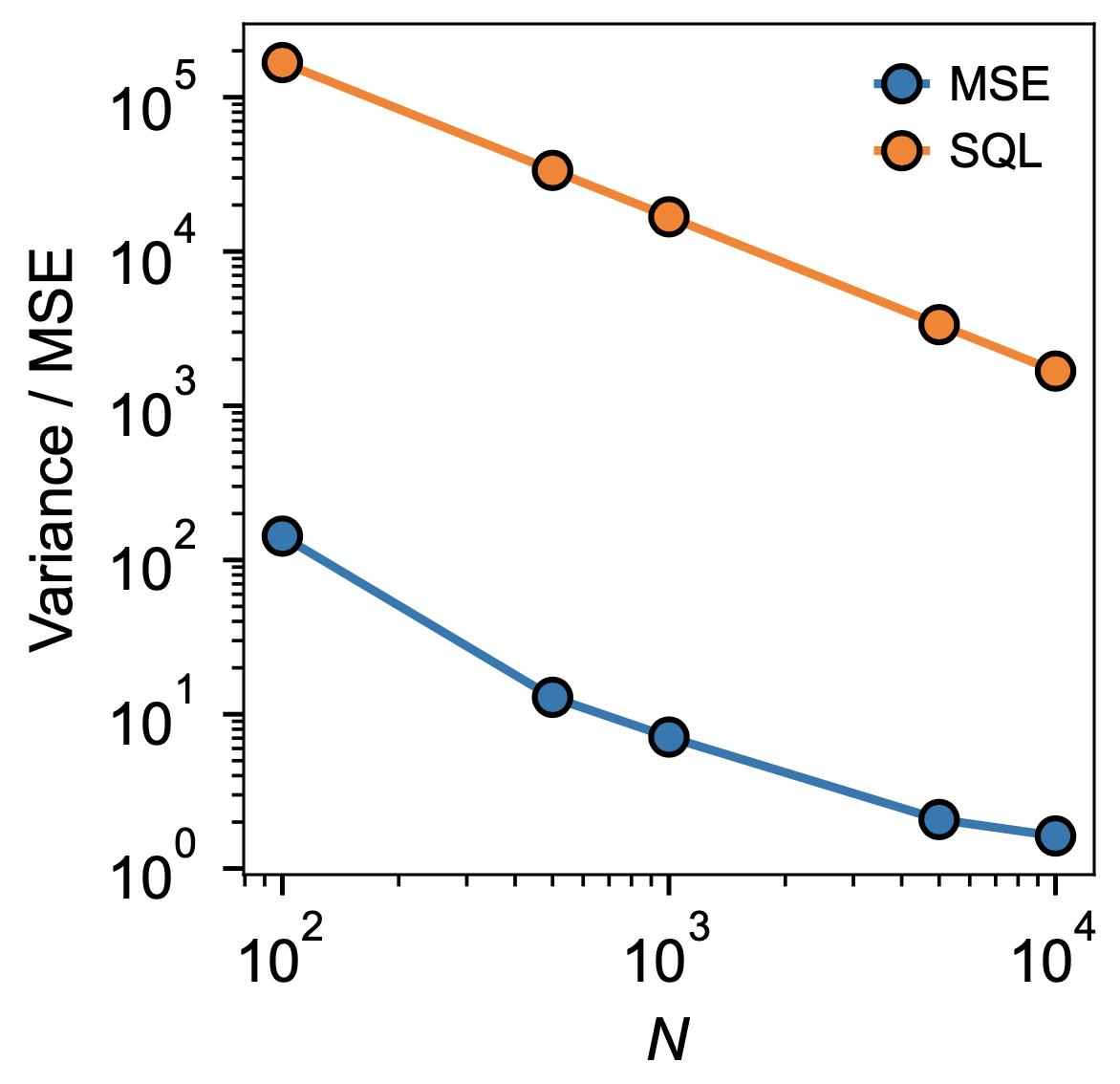}
    \caption{
    Model MSE and SQL versus $\bar{N}$ for models trained on images of triple linear sinusoids with $\bar{N}$ ranges of 40 - 4000. Here, the SQL is computed as $\sum_{x,y} 1/N(x,y)$), by assuming the pixels are uncorrelated.
    }
    \label{fig:mse_vs_SQL_uncorrelatd}
\end{figure*}

\end{document}